\def\year{2019}\relax
\documentclass[letterpaper]{article} 
\usepackage{aaai19}  
\usepackage{times}  
\usepackage{helvet}  
\usepackage{courier}  
\usepackage{url}  
\usepackage{graphicx}  
\usepackage{algorithm}
\usepackage{algorithmic}
\usepackage{amssymb}
\usepackage{amsmath}
\usepackage{amsthm}
\usepackage{subfigure}
\DeclareMathOperator*{\argmax}{arg\,max}
\DeclareMathOperator*{\argmin}{arg\,min}

\title{Active Generative Adversarial Network for Image Classification}
\author{\\ {\bf Quan Kong \thanks{contribute equally to this paper.}\footnotemark[2]} \quad {\bf Bin Tong \footnotemark[1]\footnotemark[2]} \quad {\bf Martin Klinkigt \footnotemark[2]} \quad \\ {\bf Yuki Watanabe \footnotemark[3]} \quad {\bf Naoto Akira \footnotemark[2]} \quad {\bf Tomokazu Murakami \footnotemark[2]} \\
\footnotemark[2]Hitachi, Ltd. R\&D Group, Japan\\
\footnotemark[3]Hitachi America, Ltd. USA \\
\{quan.kong.xz, bin.tong.hh, martin.klinkigt.ut\}@hitachi.com\\
\{naoto.akira.vu, tomokazu.murakami.xr\}@hitachi.com\\
yuki.watanabe@hal.hitachi.com
}

\begin{document}
\maketitle
\begin{abstract}
Sufficient supervised information is crucial for any machine learning models to boost performance. However, labeling data is expensive and sometimes difficult to obtain. Active learning is an approach to acquire annotations for data from a human oracle by selecting informative samples with a high probability to enhance performance. In recent emerging studies, a generative adversarial network (GAN) has been integrated with active learning to generate good candidates to be presented to the oracle. In this paper, we propose a novel model that is able to obtain labels for data in a cheaper manner without the need to query an oracle. In the model, a novel reward for each sample is devised to measure the degree of \textit{uncertainty}, which is obtained from a classifier trained with existing labeled data. This reward is used to guide a conditional GAN to generate informative samples with a higher probability for a certain label. With extensive evaluations, we have confirmed the effectiveness of the model, showing that the generated samples are capable of improving the classification performance in popular image classification tasks.
\end{abstract}

\section{Introduction}

Machine learning models including traditional ones and new emerging deep neural networks require sufficient supervised information, i.e., class labels, to achieve fair performance. In situations in which labeled data is expensive or difficult to obtain, these models degenerate in performance. Active learning \cite{ref:settles2009active} is proposed for handling such a problem. It aims to find the best approach to leverage a limited number of labeled data and to reduce the cost of data annotation. Active learning selects informative samples from a pool of unlabeled data and obtains their labels by involving a human oracle. In this paper, we investigate the problem of \textit{lack of labeled data} from a new and different perspective. We propose a model to improve learning performance, which is able to make use of limited labeled data without using any additional unlabeled data nor involving any human oracle to acquire labels. 

As to a classification model, informative samples are those that are able to better contribute to improving classification performance than other samples. For example, samples close to the hyper-plane are often \textit{uncertain} for a support vector machine (SVM) based classifier. Therefore, acquiring labels of those samples can reduce the \textit{uncertainty}, thereby reducing classification errors. In the area of active learning, informative samples are selected from a pool of unlabeled data by using criteria, such as degree of \textit{uncertainty}. The labels of the selected samples are obtained by querying a human oracle. Recently, there have been attempts \cite{ref:GAAL,ref:ActiveDecisionDGM} which label informative samples generated from a generative adversarial network (GAN) \cite{GoodfellowPMXWOCB14}. In these works, GAN is used to generate samples with the same distribution as the unlabeled dataset. In \cite{ref:GAAL}, latent variables, which are able to generate samples that have small distances to the classification hyper-plane, are selected. These latent variables are used to generate samples that are labeled by involving an oracle. In \cite{ref:ActiveDecisionDGM}, a GAN is used to generate samples compound of two classes and the human oracle has to choose the sample which cannot be clearly assigned to either class. However, the above methods still need to use a pool of unlabeled data and query the human oracle.

Unlike the above methods, we investigate the problem of how to acquire labeled data to improve the classification performance by only using a limited number of labeled data. A straightforward way is to use conditional GAN \cite{ref:conGAN,ref:ACGAN} for generating labeled samples. However, for a class of samples, most generated samples may fall inside its convex hull. Samples inside this convex hull are less discriminative to other classes, while samples along or even outside of the convex hull are informative to optimize the hyper-plane of the classifier. In the idea of active learning, a classifier trained with existing labeled data provides a signal to determine if a sample is uncertain to the hyper-plane of the classifier. In this work, we use this external signal to guide the conditional GAN in generating informative labeled samples with a higher probability that contribute to improving classification performance. This can be regarded as an optimization with a trade-off between generation of samples with the same distribution as the training samples and generation of informative samples. This discipline is widely used in machine learning, such as penalizing complexity of parameters to avoid over-fitting. The contribution of our work is two-fold:
\begin{enumerate}
  \item We propose a model that provides a cheaper way than active learning to acquire labeled samples. Instead of querying the oracle, our model generates labeled samples with a higher probability that are informative to optimize the hyper-plane of a classifier.  
  \item We propose a novel loss function for training generative network model to generate informative samples with a specific label that is inspired by the idea of policy gradient \cite{ref:NIPS1999_1713} in reinforcement learning. We regard generated samples and the external signal related to \textit{uncertainty} as \textit{action} and \textit{reward}, and use this reward to update the parameters of network for generating informative samples. 
\end{enumerate}

\section{Related Work}


Relying only a limited amount of labeled data is a long-standing and important problem in the area of machine learning. Different philosophies and problem settings exist for dealing with this, such as transfer learning \cite{ref:transferlearning}, semi-supervised learning \cite{ref:semisuper} and active learning \cite{ref:settles2009active}. 

Transfer learning focuses on how to optimize a model with a limited amount of labeled data by transferring the knowledge from a similar yet different source task with sufficient labeled data, thereby reducing the cost of data annotation for the task at hand. Zero-shot learning \cite{ref:zrl,ref:devise} is a variation of transfer learning, in which \textit{unseen}, and therefore unlabeled objects are expected to be recognized by transferring knowledge from \textit{seen} classes \cite{ref:settles2009active}. Semi-supervised learning leverages the unlabeled data to boost the performance in case of only labeled data is used. Active learning \cite{ref:settles2009active} is based on a different philosophy. Typically, unlabeled data is available to this learning paradigm, in which the most informative samples from the pool of unlabeled data are selected to query an oracle. Active learning provides a schema to limit the number of queries by selecting the most informative samples to maximize the effect of the acquired labels. To determine the degree of \textit{uncertainty} used in the query strategy, uncertainty sampling \cite{ref:5206651} is the most simple, yet widely used criterion to measure informativeness. Other criteria for the query strategy may include query by committee (QBC) \cite{ref:Freund1997}, expected error reduction \cite{ref:Roy01towardoptimal,ref:10.1007/978-3-540-74565-5_47} and density weighted methods \cite{ref:shen-EtAl:2004:ACL,ref:10.1007/978-3-540-74958-5_14,ref:Cebron2009}.

A generative adversarial network (GAN) \cite{GoodfellowPMXWOCB14} is a neural network model trained in an unsupervised manner, aiming to generate new data with the same distribution as the data of interest. It is widely applied in computer vision and natural language processing tasks, such as generating samples of images \cite{ref:gan_gen_image} and generating sequential words \cite{ref:gan_dialog}. One of its variants, conditional GAN \cite{ref:conGAN}, uses both label information and noisy latent variables to generate samples for a specified label. A variant of conditional GAN, called Auxiliary Classifier GAN (AC-GAN) \cite{ref:ACGAN}, uses supervised information to generate high quality images at pixel level. 

Recently, the GAN models have been used with transfer learning \cite{ref:transferlowshot,ref:UnsupervisedPixel}, zero-shot learning \cite{ref:binZero,ref:zrl_ccdeepgen}, semi-supervised learning \cite{ref:Dai2017GoodSL,ref:lee2018training} and active learning \cite{ref:GAAL,ref:ActiveDecisionDGM}. In \cite{ref:Dai2017GoodSL} and \cite{ref:lee2018training}, GAN models are used to train feed-forward classifiers with an additional penalty derived from out-of-distribution samples or low-density samples. It should also be emphasized that both low-density and out-of-distribution samples does not necessarily represent hard samples. For example, out-of-distribution samples, which are far from the classification boundary, will not be regarded as hard samples. Another difference from them lies in that our work focuses on training the generator with an informativeness reward given by the existing classifier. Unlike the works \cite{ref:GAAL,ref:ActiveDecisionDGM} in which the generated samples are presented to the oracle, this work focuses on directly generating informative labeled samples that might contribute to boosting learning performance. To the best of our knowledge, this is the first study that uses a GAN to generate informative samples by incorporating a new devised factor to measure the degree of \textit{uncertainty}. 
This study provides a new paradigm that augments labeled data to improve learning performance without using any other unlabeled data nor involving a human oracle.

\begin{figure*}[ht]
\centering
\includegraphics[scale=0.420]{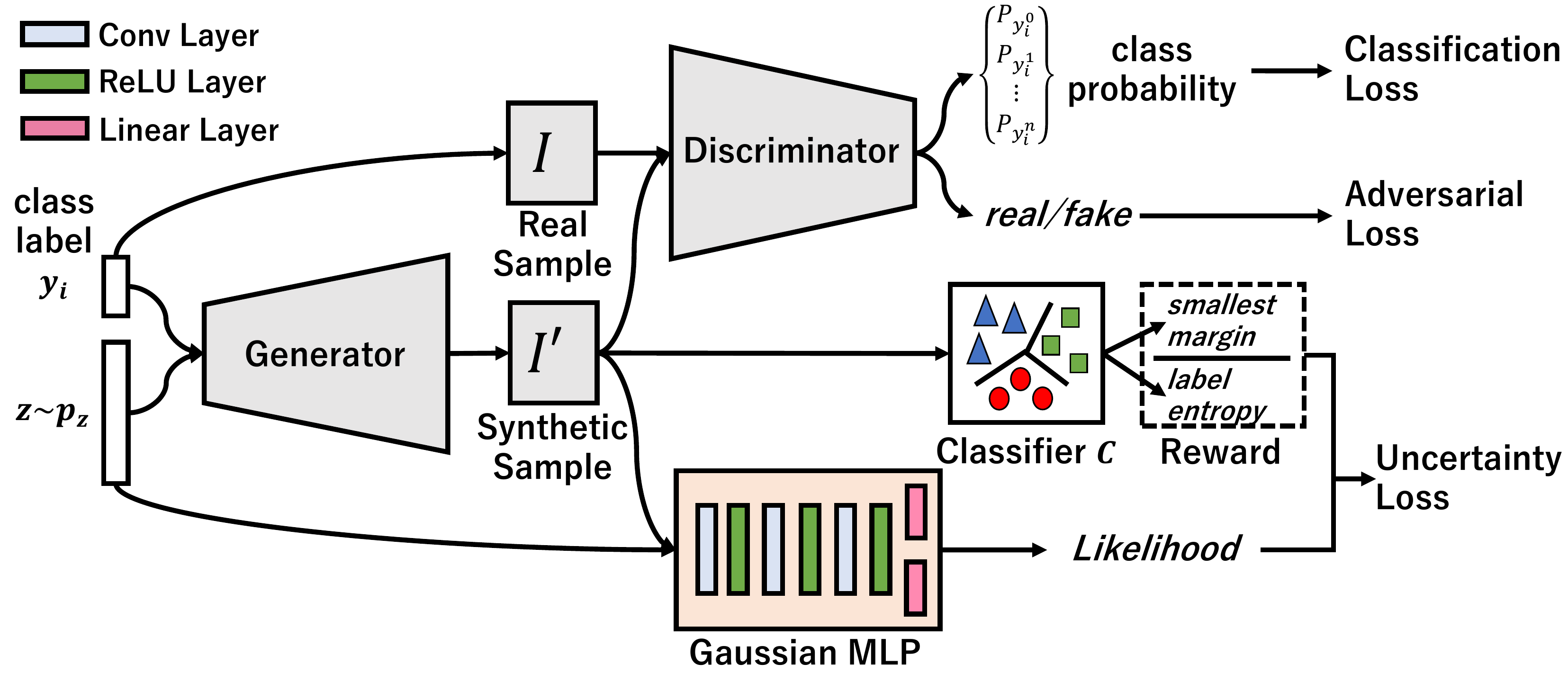}
\caption{Architecture of our proposed model. \textbf{I.} The generator generates samples by training a generator and a discriminator with a class label one hot vector $y_i$ from training sample as a condition. In our case, we introduce a Classification Loss follows AC-GAN \cite{ref:ACGAN} to make sure the generated sample is highly related to the specific given label condition. The discriminator also determines the possibility if a generated sample is fake, which is used as our Adversarial Loss to make the distributions of generated and real samples similar. \textbf{II.} Simultaneously, a classifier $C$ trained with existing labeled data. It calculates a reward for a sample related to the degree of \textit{uncertainty} includes the smallest margin and label entropy. \textbf{III.} A Gaussian Multi-Layer Perceptron (MLP) \cite{ref:Duan2016} calculates a likelihood of a sample to be generated from current distribution of synthetic samples. Uncertainly Loss consists of a likelihood and rewards of each generated sample, and we use it to train the generator for generating informative samples via policy gradient.}
\label{fig:overview}
\end{figure*}

\section{Preliminary}

In this section, we introduce the preliminaries of GAN and active learning that serves as the basis to derive our new model. A GAN \cite{GoodfellowPMXWOCB14} consists of a generator $G$ and discriminator $D$ that compete in a turn-wise min-max game. The discriminator attempts to distinguish real samples from synthetic samples, and the generator attempts to fool the discriminator by generating synthetic samples looking like real samples. The $D$ and $G$ play the following game on $V(D, G)$
\begin{eqnarray}
\label{eq_gan}
\min_{G}\max_{D} V(D, G) = \mathbb{E}_{\mathbf{x}_i\in p_{data}(\mathbf{x})}[\log D(\mathbf{x}_i)] + \nonumber \\ \mathbb{E}_{z \in p_{z}(z)}[\log (1-D(G(z)))],
\end{eqnarray}
where $\mathbf{x}_i$ represents a sample. $p_{data}$ and $p_{z}$ represent the distribution of real samples and synthetic samples, and $z$ represents a noise vector.
A GAN tries to map $p_{z}$ to $p_{data}$; that means the generated samples from $G$ are desired to own a high likeness with $p_{data}$ which is also the distribution of training samples. In the original GAN model only $z$ is used to generate samples. In a variation called conditional GAN (CGAN) \cite{ref:conGAN}, a condition $y_i$, which is a class label of $\mathbf{x}_i$, is included in addition to $z$ to control the sample generation. The objective function becomes
\begin{eqnarray}
\label{eq_condition_gan}
\min_{G}\max_{D} V(D, G) = \mathbb{E}_{\mathbf{x}_i\in p_{data}(\mathbf{x})}[\log D(\mathbf{x}_i|y_i)] + \nonumber \\ \mathbb{E}_{z \in p_{z}(z)}[\log (1-D(G(z|y_i)))],
\end{eqnarray}
where $y_i$ could be a one-hot representation of the class label. During training of the CGAN model, $y_i$ is used to instruct the generator $G$ to synthesize samples for this given class.

Active learning is a machine learning method that is able to interactively query an oracle to obtain labels for samples. These samples are selected from an unlabeled sample pool by using a criterion to measure if the selected sample is able to reduce the learning error. To be more specific, standard supervised learning problems assume an instance space of data $X$ and labels $Y$. A mapping function $f:X \rightarrow Y$ is optimized by minimizing error:
\begin{eqnarray}
f^{\ast} = \argmin_{f \in F}\sum_{Y}L(f(X),Y), 
\end{eqnarray}
where $F$ represents a space over a predefined class of functions. The error is measured by a loss function $L$ that penalizes disagreement between $f(X)$ and $Y$. In the typical setting of active learning \cite{settles2009active}, a pool of unlabeled samples $U = \{\mathbf{x}_1^u, \dots, \mathbf{x}_n^u\}$ is given. Denote $M = \{(\mathbf{x}_1,y_1), \dots, (\mathbf{x}_n,y_n)\}$, where $\mathbf{x}_i \in X$ and $y_i \in Y$. Active learning performs in an iterative way: (1) training a classifier $f$ on $M$; (2) using a query function $Q(f, M, U)$ to select an unlabeled sample $i^{\ast}$ to label; (3) removing $\mathbf{x}_{i^{\ast}}^u$ from $U$ and adding $(\mathbf{x}_{i^{\ast}}^u, y_{i^{\ast}})$ to $M$. The target of active learning is to choose samples $i^{\ast}$ to be labeled by asking an oracle, and reduce the learning error with as few queries as possible. The selected samples are regarded as more informative than other unselected ones in terms of contribution in learning error reduction.

\section{Proposed Method}

In this section, we discuss the details of the proposed method. Without loss of generality, we take a classifier as the example of supervised learning, in which we are given a set of labeled data $\mathcal{S}_l = \{(\mathbf{x}_1, y_1), (\mathbf{x}_2, y_2), \dots, (\mathbf{x}_N, y_N)\}$ where $\mathbf{x}_i$ is a sample, $y_i$ is its corresponding label, and $N$ represents the number of samples.

The overview of our proposed model is shown in Figure \ref{fig:overview}. This model mainly consists of a classifier $C$ trained with existing labeled data, a conditional GAN, and a Gaussian Multi-Layer Perceptron (MLP)  \cite{ref:Duan2016}. The conditional GAN is used to generate labeled samples. A novel reward is devised to measure the degree of informativeness for each generated sample. This reward is calculated according to a degree of \textit{uncertainty} for a sample with respect to the hyper-plane of the pre-trained classifier. In general, the more informative a sample is, the higher probability this sample is able to improve classification performance if it is included in the existing labeled data. The Gaussian MLP provides a likelihood of a generated sample to be generated from a recent set of generated samples. Together with the likelihood, the reward is used to update parameters for the generator in the conditional GAN. This model makes a trade-off between generating samples with the same distribution as the labeled data and generating informative samples to improve classification performance for the pre-trained classifier.

\subsection{Generation of labeled samples}

Since we focus on image classification tasks, the proposed model uses a variant of conditional GAN, called AC-GAN \cite{ref:ACGAN}, which shows its promising performance on generating images.

Given a set of labeled images $\{(\mathbf{x}_1,y_1), \dots, (\mathbf{x}_N,y_N)\}$, the AC-GAN model is used to generate labeled samples with the inputs of both a noise latent variable and a one-hot representation of a class label. In the AC-GAN, the generator $G$ generates a synthetic sample $\mathbf{\widehat{x}}_{i} = G(z,y_i)$ with the noise latent vector $z$ and a label $y_i$. The discriminator gives two kinds of probabilities. One is a probability distribution over sources, i.e., $P(\text{real}|\mathbf{x}_i)$ and $P(\text{fake}|\mathbf{\widehat{x}}_i)$. The other one is posterior probabilities over the class labels, i.e., $P(y_i| \mathbf{x}_i)$ and $P(y_i| \mathbf{\widehat{x}}_i)$. The objective functions of generator and discriminator in the AC-GAN are formulated as 
\begin{eqnarray}
\begin{split}
\label{eq:acgan_discriminator}
L^{D}_{\text{AC-GAN}} & = \mathbb{E}[\log P(\text{real}|\mathbf{x}_i)] + \mathbb{E}[\log P(\text{fake}|\mathbf{\widehat{x}}_i)] + \\
 & \hspace{0.48cm} \mathbb{E}[\log P(y_i| \mathbf{x}_i)] + \mathbb{E}[\log P(y_i| \mathbf{\widehat{x}}_i)]
\end{split}
\end{eqnarray}
\begin{eqnarray}
L^{G}_{\text{AC-GAN}} = \mathbb{E}[\log P(\text{real}|\mathbf{\widehat{x}}_i)] + \mathbb{E}[\log P(y_i| \mathbf{\widehat{x}}_i)].
\end{eqnarray}
The discriminator $D$ is trained to maximize $L^{D}_{\text{AC-GAN}}$, and the generator $G$ is trained to maximize $L^{G}_{\text{AC-GAN}}$. For the discriminator $D$, the first two terms in Equation \ref{eq:acgan_discriminator} encourage that both real and fake samples are classified correctly. The last two terms in Equation \ref{eq:acgan_discriminator} encourage that both real and fake samples have correct class labels. For the generator $G$, it is expected that generated samples are classified as fake, and have correct class labels as well. 

\subsection{Measure of uncertainty}

In this subsection, we discuss how the degree of \textit{uncertainty} is measured in the proposed model. Among the samples generated by the AC-GAN model, only informative samples might be able to contribute to improving classification performance. In the area of active learning, uncertainty sampling is the most widely used query strategy. The intuition behind uncertainty sampling is that if a sample is highly uncertain with a hyper-plane of a classifier, obtaining its label will improve the degree of discrimination among classes. In other words, this sample is considered to be informative in improving the classification performance. In our model, we use SVM as the classifier. In our paper, we mainly use two metrics based on the label probabilities to measure the uncertainty of a sample.

\textbf{Smallest Margin} Margin sampling is an uncertainty sampling method in the case of multi-class \cite{ref:settles2009active}, which is defined as
\begin{eqnarray}
\label{eq:smallest_margin}
\mathbf{\widehat{x}}_M = \argmin_{\mathbf{\widehat{x}}_i}(P(y'_1|\mathbf{\widehat{x}}_i) - P(y'_2|\mathbf{\widehat{x}}_i)),
\end{eqnarray}
where $y'_1$ and $y'_2$ are the first and second most probable class labels of a generated sample $\mathbf{\widehat{x}}_i$ under the specified classifier, respectively. Intuitively, samples with large margins are easy, since the classifier has little doubt in differentiating between the two most likely class labels. Samples with small margins are more ambiguous, thus knowing the true label will help the model to discriminate more effectively between them.

\textbf{Label Entropy} A more general uncertainty sampling strategy uses the entropy of posterior probabilities over class labels. In smallest margin, posterior probabilities of labels other than the two most probable class labels are simply ignored. To mitigate this problem, the entropy over all class labels is used, which is formulated as
\begin{eqnarray}
\label{eq:entropy_label}
\mathbf{\widehat{x}}_{LE} = \argmax_{\mathbf{\widehat{x}}_i}-\sum_{y'} p(y'|\mathbf{\widehat{x}}_i) \log p(y'|\mathbf{\widehat{x}}_i).
\end{eqnarray}

\subsection{Loss on uncertainty}

In this subsection, we discuss how to devise a loss function for the generated samples based on the degree of \textit{uncertainty} to update the parameters of the generator.

Policy gradient \cite{ref:NIPS1999_1713} has been successfully applied in reinforcement learning to learn an optimal policy. As one target of this work is to guide the generator to synthesize informative samples, we regard the degree of \textit{uncertainty} and the generated samples as \textit{reward} and \textit{action}, respectively. In general, the higher the degree of \textit{uncertainty} is, the higher the reward is obtained. If a generated sample has a high degree of \textit{uncertainty}, this sample is encouraged to be generated with a high probability. To the best of our knowledge, we are the first to use the idea from policy gradient to model the degree of \textit{uncertainty} in active learning.

In the following we discuss how to convert the degree of \textit{uncertainty} into a reward. With respect to smallest margin, for each generated sample $\mathbf{\widehat{x}}_i$, the reward can be simply calculated by $\mathbf{r}_m(\mathbf{\widehat{x}}_i) = e^{-u_m}$, where $u_m = P(y'_1|\mathbf{\widehat{x}}_i) - P(y'_2|\mathbf{\widehat{x}}_i)$. If the difference between the probabilities of the two most probable class labels for a generated sample is small, this generated sample is uncertain. This results in a larger value of $\mathbf{r}_m$ than other certain samples. Based on the nature of Equation \ref{eq:smallest_margin}, $u_m$ falls into the range of $[0, 1]$. The values of $\mathbf{r}_m$, which fall in the range $[\frac{1}{e}, 1]$, have no significant difference between the best and worst cases, which may result in an inappropriate design of the reward. Inspired by \cite{ref:lee2018training}, we set a threshold $\epsilon$ to truncate the value of reward for a bad case where the margin of two probabilities is large. Specifically, given a threshold $\epsilon$, the reward is 
\begin{eqnarray}
\label{eq:margin_reward}
    \mathbf{r}_m(\mathbf{\widehat{x}}_i)= 
\begin{cases}
    e^{-u_m},& \text{if } u_m\leq \epsilon\\
    C,              & \text{otherwise}
\end{cases}
\end{eqnarray}
where $C$ is a constant number. In our work, we set its value to $0$, which means that if $u_m$ is larger than $\epsilon$, we enforce the reward $\mathbf{r}_m$ to be zero. 

With respect to label entropy, we can calculate the reward similar to $\mathbf{r}_{le}(\mathbf{\widehat{x}}_i) = e^{u_{le}}$, where $u_{le} = -\sum_{y'} p(y'|\mathbf{\widehat{x}}_i) \log p(y'|\mathbf{\widehat{x}}_i)$. The reward for a generated sample is calculated by combining the above two factors, which is formulated as
\begin{eqnarray}
\label{eq:reward}
\mathbf{r}(\mathbf{\widehat{x}}_i) = \alpha \cdot \mathbf{r}_m(\mathbf{\widehat{x}}_i) + (1 - \alpha) \cdot \mathbf{r}_{le}(\mathbf{\widehat{x}}_i),
\end{eqnarray}
where $\alpha$ is a parameter that balances the importance between the two metrics of smallest margin and label entropy. 
According to policy gradient, we devise the loss for generated samples formulated as follows:
\begin{eqnarray}
\label{eq:activeloss}
L_{\text{uncertainty}} = \sum_{\mathbf{\widehat{x}}_i}  \mathbf{r}(\mathbf{\widehat{x}}_i) P(\mathbf{\widehat{x}}_i|\theta),
\end{eqnarray}
where $\mathbf{\widehat{x}}_i$ represents a generated sample from the generator $G(z, y_i)$, $P(\mathbf{\widehat{x}}_i|\theta)$ represents the probability of $\mathbf{\widehat{x}}_i$ that is generated by the generator. However, the generator does not directly provide such a probability for each generated sample. Therefore, we have to estimate this probability based on a model with the parameters $\theta$. In our work, we choose a MLP to parameterize the policy. We use a Gaussian distribution over action space, where the covariance matrix was diagonal and independent of the state. The Gaussian MLP maps from the input synthetic image $G(z,y_i)$ to the mean $\mathbf{\mu}$ and standard deviation $\mathbf{\sigma}$ of a Gaussian distribution with the same dimension as $z$. Thus, the policy is defined by the normal distribution $\mathcal{N}(\theta|\mathbf{\mu}, e^\mathbf{\sigma})$. Then we can compute the likelihood $P(G(z,y_i)|\theta)$ with $\mathbf{\mu}$ and $\mathbf{\sigma}$ from the output of approximated Gaussian MLP. The Gaussian MLP is jointly learned with $G$ and $D$ by policy gradient.

\begin{algorithm}[ht]
\caption{ActiveGAN} \label{alg:activeGAN}

\hspace*{\algorithmicindent} \textbf{Input} training data $\mathbf{x}_i$ and its label $y_i$ where $i \in [1, \dots, N]$.\\
\hspace*{\algorithmicindent} \textbf{Output} $\Psi_d$ (parameters of D), $\Psi_g$ (parameters of G) and $\theta$ (parameters of MLP)
\begin{algorithmic}[1]

\STATE Initialize $\alpha$, $\lambda$, $\theta$, $\Psi_d$ and $\Psi_g$.
\STATE Set the buffer size to be $M$
\STATE Train SVM with grid-search for best parameters 
\STATE Train the generator $G$ and the discriminator $D$ with first $m$ iterations
\STATE Save generated samples in $m$ iterations into the buffer
\REPEAT
\STATE Generate a sample $\mathbf{\widehat{x}}_i$ $\leftarrow$ $G(z, y_i)$ 
\STATE Use Equation \ref{eq:reward} to calculate the reward $\mathbf{r}(\mathbf{\widehat{x}}_i)$ for  $\mathbf{\widehat{x}}_i$.
\STATE Use generated samples to calculate the likelihood $P(\mathbf{\widehat{x}}_i|\theta)$ for $\mathbf{\widehat{x}}_i$
\STATE Use Equation \ref{eq:activeloss} to calculate the loss $L_U$ related to the degree of $uncertainty$ for $\mathbf{\widehat{x}}_i$
\STATE Update parameters for the generator $G$ and $MLP$: $\Psi_g$,$\theta$ $\leftarrow$ ($\Psi_g$,$\theta$) + $\bigtriangledown_{\Psi_g,\theta}$ $L^{G}_{\text{ActiveGAN}}(\Psi_g,\theta)$ 
\STATE Update parameters for the discriminator $D$: $\Psi_d$ $\leftarrow$ $\Psi_d$ + $\bigtriangledown_{\Psi_d}$ $L^{D}_{\text{AC-GAN}}$
\STATE Update the buffer by adding the sample $\mathbf{\widehat{x}}_i$
\UNTIL
\end{algorithmic}
\end{algorithm}

\subsection{Algorithm}
\label{sec:algorithm}

By integrating the loss measuring the degree of \textit{uncertainty} for the generated samples, our proposed model, called ActiveGAN, has the following loss function for the generator, which is maximized.
\begin{flalign}
\label{eq:ActiveGAN_gen}
L^{G}_{\text{ActiveGAN}} &= L^{G}_{\text{AC-GAN}} + \lambda L_{\text{uncertainty}} \nonumber &&\\ 
&=\nonumber \mathbb{E}[\log P(\text{real}|\mathbf{\widehat{x}}_i)] + \mathbb{E}[\log P(y| \mathbf{\widehat{x}}_i)] &&\\
& \hspace{0.32cm} + \lambda \mathbb{E}[P(\mathbf{\widehat{x}}_i|\theta) \mathbf{r}(\mathbf{\widehat{x}}_i)],
\end{flalign}
where $L^{G}_{\text{AC-GAN}}$ is the loss function of the generator in AC-GAN. The discriminator in \text{Active-GAN} is the same as that in \text{AC-GAN}, which is denoted by $L^{D}_{\text{ActiveGAN}}$. The notations $\Psi_g$ and $\Psi_d$ in Algorithm \ref{alg:activeGAN} represent the parameters of generator and discriminator in the \text{ActiveGAN}, respectively. $\lambda$ is a parameter that balances the importance between the loss for the generator in the AC-GAN model and the loss related to the degree of \textit{uncertainty} for the generated samples. The larger the value of $\lambda$ is, the more likely the model is forced to generate samples that contribute to improving the classification performance instead of generating samples whose distribution is the same as the training ones. The learning process of ActiveGAN is depicted in detail in Algorithm \ref{alg:activeGAN}. Following \cite{ref:NIPS1999_1713}, the gradient of the objective function $L_{\text{uncertainty}}$ can be derived as
\small 
\begin{flalign}
\label{eq:policy_gradient}
\bigtriangledown_{\Psi_g,\theta}L_{\text{uncertainty}} = \mathbb{E}[\bigtriangledown_{\Psi_g,\theta}\log P(G_{\Psi_g} (z,y_i)|\theta) \mathbf{r}(\mathbf{\widehat{x}}_i)]
\end{flalign}
\normalsize

The evaluation of ActiveGAN is conducted as follows. We use the trained generator $G$ to synthesize a specific number of samples, which we denoted by $\mathcal{S}_g$. Together with the labeled data $\mathcal{S}_l$, we retrain the SVM to examine if improvement of classification performance is achieved. 

\section{Experiments}
In this section, we introduce evaluation settings and discuss performances of models. We then make analysis for a further understanding on our model.

\subsection{Evaluation settings}

\def\arraystretch{1.1}
\begin{table*}[t]
\centering
\caption{F-score of models on CIFAR-10, MNIST, Fashion-MNIST (F-MNIST) and Tiny-ImageNet. $n$ represents the number of labeled images used for training.}
\begin{tabular}
{l l l | l l | l l | l l}
\hline
\hline
 & \multicolumn{2}{c}{CIFAR-10} & \multicolumn{2}{c}{MNIST} & \multicolumn{2}{c}{F-MNIST} & \multicolumn{2}{c}{Tiny-ImageNet}\\
Method & $n$=5k & $n$=10k & $n$=500 & $n$=1k & $n$=5k & $n$=10k & $n$=10k & $n$=20k\\
\hline
\text{Baseline(SVM)} & 83.4  & 85.3  & 94.6  & 96.2  & 87.1  & 88.1  & 56.1  & 58.3  \\
\hline
\text{AC-GAN} & 81.4  & 82.7  & 94.1  & 95.8  & 85.4  & 86.4  & 52.2  & 56.1  \\
\text{AC-GAN+F} & 82.5  & 83.2  & 94.5  & 95.9  & 86.2  & 87.3  & 53.2  & 56.9  \\
\hline
\text{ActiveGAN} & \textbf{84.3} & \textbf{86.3} & \textbf{95.1} & \textbf{96.5 } & \textbf{87.6 } & \textbf{89.0 } & \textbf{57.5 } & \textbf{59.4 }\\
\hline
\hline
\end{tabular}
\label{tbl:overview_result}
\end{table*}

We utilized four datasets CIFAR10 \cite{ref:cifar10}, MNIST \cite{ref:svhn}, Fashion-MNIST \cite{ref:xiao2017_online} and a large scale dataset Tiny-ImageNet \cite{ref:Russakovsky} for evaluation of the proposed model ActiveGAN. MNIST consists of 50,000 training samples, 10,000 validation samples and 10,000 testing samples of handwritten digits of size 28 $\times$ 28. CIFAR10 has colored images for 10 general classes. Again we find 50,000 training samples and 10,000 testing samples of size 32 $\times$ 32 in CIFAR10. Fashion-MNIST has a training set of 60,000 examples and a test set of 10,000 examples. Each example is a 28 $\times$ 28 grayscale image, associated with a label from 10 different classes associated with fashion items. Tiny-ImageNet has 200 classes, each class has 500 training images, 50 validation images, and 50 test images. All images are 64 $\times$ 64.

We used the same network structure for the generator and discriminator as in \cite{ref:ACGAN} for CIFAR-10 and \cite{ref:NIPS2016_6399} for MNIST and Fashion-MNIST. We downsized Tiny-ImageNet samples from 64 $\times$ 64 to 32 $\times$ 32 and use the same network structure as CIFAR-10. To train a stable ActiveGAN, the parameters of the discriminator are updated once after those of the generator are updated for a specified number of iterations. Adam was used as the gradient method for learning parameters of the network. Its initial learning rate is searched in the set $\{0.0002, 0.001\}$. We used SVM as a base classifier and its optimal hyper-parameters are chosen via a grid search. We used a pre-trained VGG-16 \cite{ref:Simonyan14c} to extract features for images for all datasets. The threshold $\epsilon$ in Equation \ref{eq:margin_reward} was set to $0.2$. The balancing parameter $\alpha$ in Equation \ref{eq:reward} was set to $0.5$. The balancing parameter $\lambda$ in Equation \ref{eq:ActiveGAN_gen} was set to $0.1$ to guarantee that values of two terms $L^{G}_{\text{AC-GAN}}$ and $L_U$ are in the same scale. 

We compared the performance of the proposed model for a number of settings. Together with images in the training set, the generated images from \text{AC-GAN} or \text{ActiveGAN} are used to retrain the SVM. For the fair comparison, we had two different settings for dealing with those generated images for the compared method AC-GAN. The first setting is to use all generated images from \text{AC-GAN}. The second one is to use the model, denoted by \text{AC-GAN+F}, in which samples are generated from \text{AC-GAN} and an additional filter with a margin is then applied to these generated samples to obtain informative samples. This filter attempts to filter out the samples that overlap the training samples, leaving samples outside the distribution of training samples. The margin $u_m$, which is the difference of posterior probabilities of the two most probable class labels, is calculated by Equation \ref{eq:smallest_margin} and this margin was tuned in our evaluation. The F-score is used as the metric for evaluating performance. F-score is calculated by $\frac{2 \cdot P \cdot R}{P + R}$, where $P$ are $R$ are precision and recall, respectively.

\subsection{Image Classification}

Table \ref{tbl:overview_result} shows the classification performances of models for four different datasets. Note that the margin for \text{AC-GAN+F} was also tuned for each data set in the set $\{0.1,0.15,0.20,0.25,0.3\}$. The best F-scores were chosen, which are $0.2$, $0.15$, $0.15$, and $0.2$ in CIFAR-10, MNIST, F-MNIST and Tiny-ImageNet, respectively. We chose the number of training samples $n$ as $500$ or $1,000$ for MNIST and as $5,000$ or $10,000$ for the other datasets. Note that all test samples in each dataset were used. 

We can see that when the number of images used for training the baseline SVM increases, F-scores were improved in every dataset. When the generated images from \text{AC-GAN} were used for training together with the existing labeled images, the classification performance even dropped. For instance, in the dataset CIFAR-10, the classification performance dropped by $2.0$ points and $2.5$ points in the cases of $n$ set to $5,000$ and $10,000$, respectively. By applying the additional filter with the smallest margin, the classification performances of \text{AC-GAN+F} were slightly improved. However, the side effect is not fully mitigated, and performances were still lower than the baseline SVM. This empirically explains that the generated samples from \text{AC-GAN}, whose distribution was similar to that of images in the original training data, did not provide more information than what the original training images provide.

Using the generated samples from \text{ActiveGAN} is able to achieve better performance than the baseline classifier for all dataset without any additional data. For instance, for the dataset CIFAR10, \text{ActiveGAN} is able to achieve better performance than the baseline SVM by $0.9$ points and $1.0$ points. For the more complex dataset Tiny-ImageNet, \text{ActiveGAN} achieved F-scores of $57.5$ and $59.4$ in two different settings, which are improvements of $1.4$ points and $1.1$ points compared to the baseline SVM, respectively. These results confirm the superiority of our model over the baselines in both small-scale and large-scale image classification, which may bring \text{ActiveGAN} into a practical use.

\begin{figure}[ht]
\centering
\includegraphics[scale=0.325]{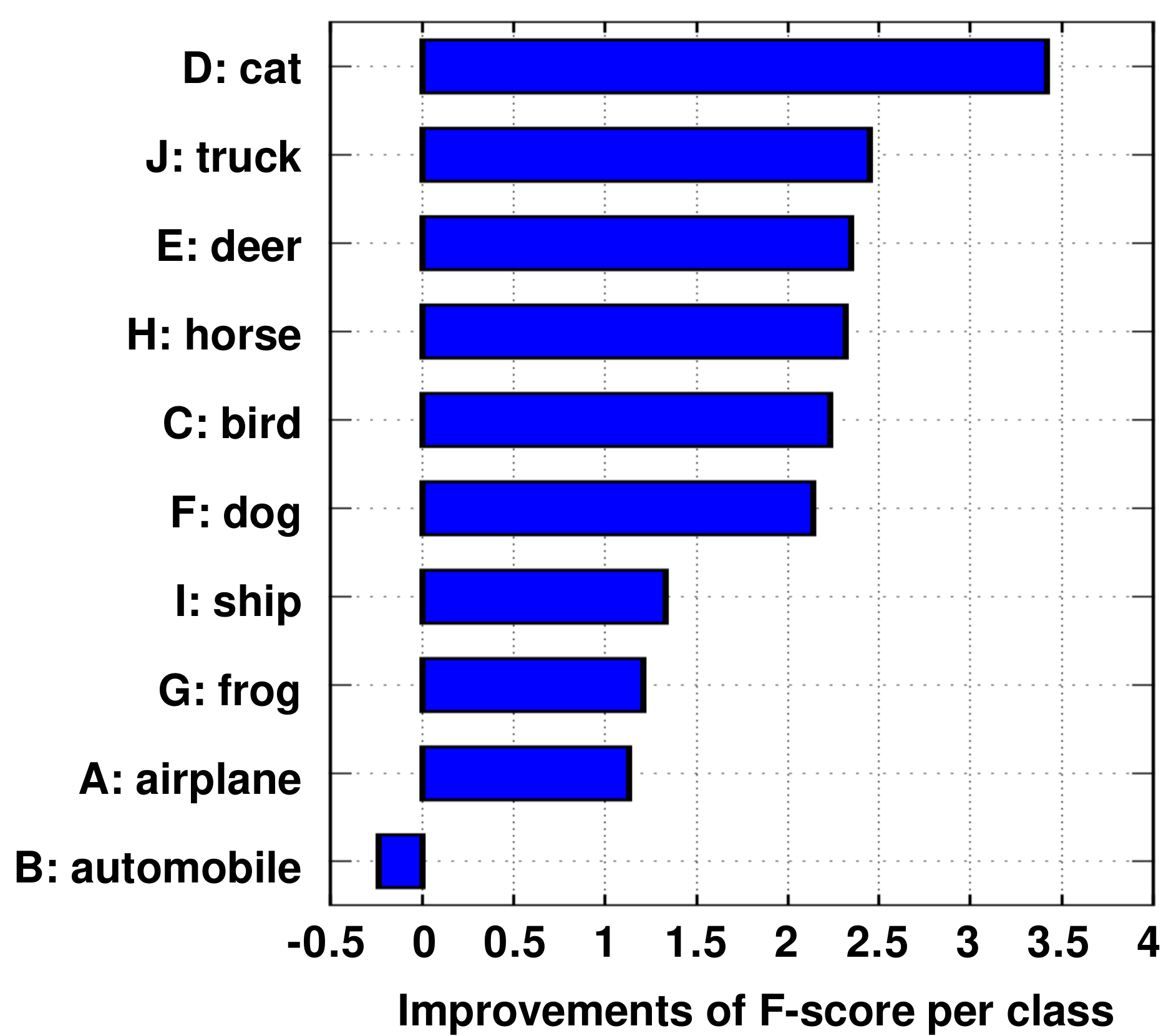}
\caption{F-score improvements of each class compared with baseline SVM in the dataset CIFAR-10. The number of images used in the training is $10$k.}
\label{fig:class_improvement}
\end{figure}

For an in-depth analysis, we also showed how the F-scores of each category changes compared to the baseline SVM if \text{ActiveGAN} is used. Figure \ref{fig:class_improvement} shows improvements of F-score for each class in the dataset CIFAR-10. We can observe that, except the class `automobile', F-scores of the other classes were improved. Among the improvements, the F-score of the class `cat' was improved by $3.42$ points. 

\subsection{Discussion and Analysis}

We examined the impact of hyper-parameter $\epsilon$ in Equation \ref{eq:margin_reward} by using CIFAR10 data. The result is shown in Table \ref{tbl:ablation_study_margin}. Note that the value of $\mathbf{r}_m(\mathbf{\widehat{x}}_i)$  without a truncation of the threshold $\epsilon$ falls into the range $[\frac{1}{e}, 1]$. The value of $\mathbf{r}_m(\mathbf{\widehat{x}}_i)$ smaller than about $0.5$ implies that the generated samples are somewhat certain to the classifier. This will put little penalty on the generated samples, thereby resulting in an inappropriate reward. Therefore, we tuned the value of $\epsilon$ from $0.0$ to $0.6$. We can see that the best performance was achieved when $\epsilon$ is set to $0.2$. Setting $\epsilon$ to $1.0$ is an extreme case. It means that the reward from smallest margin is degraded to $\mathbf{r}_m(\mathbf{\widehat{x}}_i) = e^{-u_m}$. Its performance is the worst among the other settings of $\epsilon$. It is because the generated samples, which are certain to the classifier, receive little penalty.

\renewcommand{\arraystretch}{1.1}
\setlength{\tabcolsep}{3pt}
\begin{table}[h]
\centering
\caption{F-scores when $\epsilon$ changes for $n$=10k.}
\begin{tabular}{l l l l l l l l l}
\hline
\hline
$\epsilon$ & \multicolumn{1}{c}{0.0} & \multicolumn{1}{c}{0.1} & \multicolumn{1}{c}{0.2} & \multicolumn{1}{c}{0.3} & \multicolumn{1}{c}{0.4} & \multicolumn{1}{c}{0.5} & \multicolumn{1}{c}{0.6} & \multicolumn{1}{c}{1.0}\\
\hline
F-score & 85.4 & 85.7 & 86.3 & 85.8 & 85.3 & 85.2 & 85.1 & 85.0\\
\hline
\hline
\end{tabular}
\label{tbl:ablation_study_margin}
\end{table}

\begin{figure}[!htp]
\centering
\includegraphics[scale=0.28]{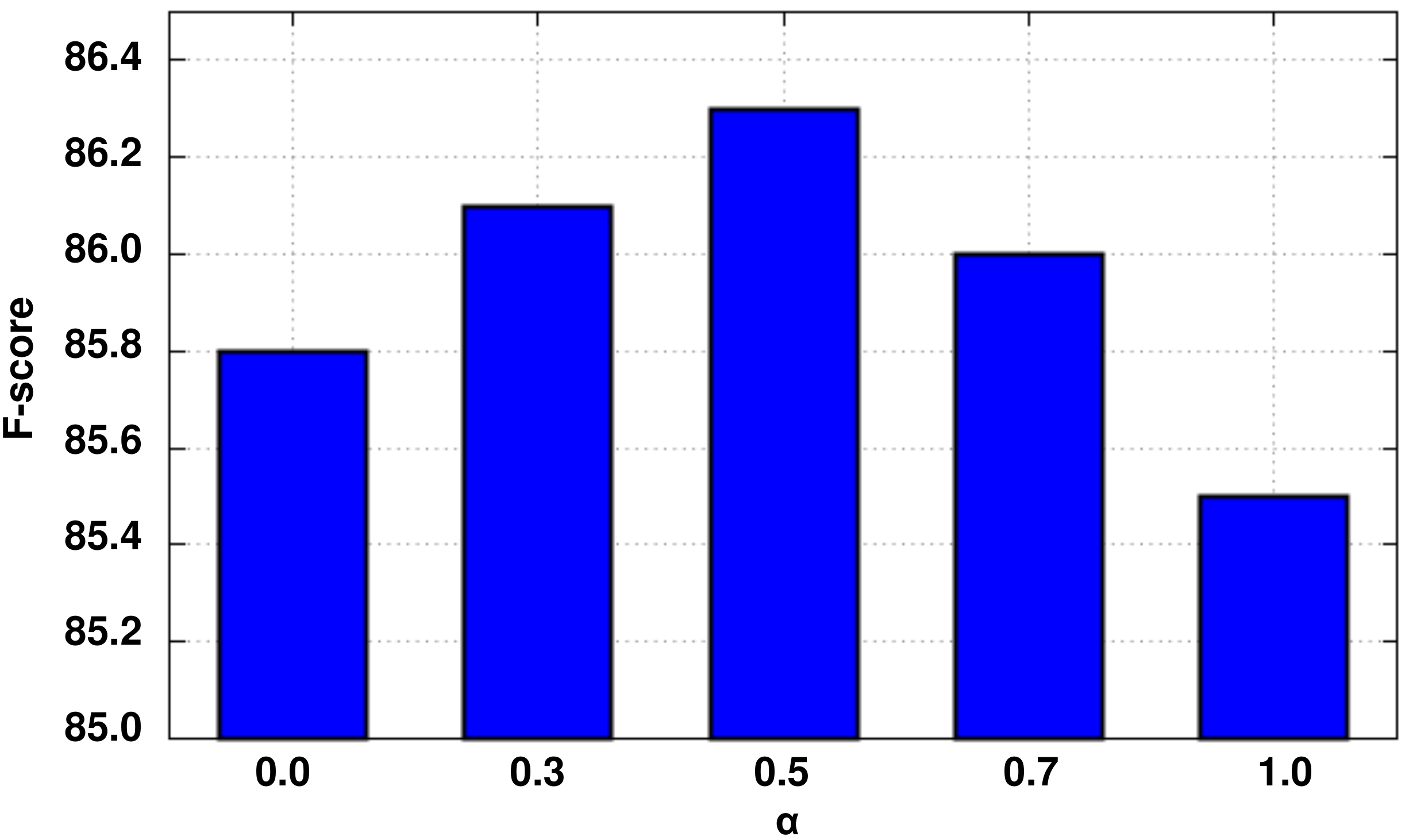}
\caption{F-scores when $\alpha$ changes for n=10k.}
\label{fig:alpha_ablation}
\end{figure}

In order to verify the effectiveness of reward derived from both smallest margin and label entropy in generating informative samples, we showed the impact of $\alpha$ in Equation \ref{eq:reward} by using CIFAR-10 data. The result is shown in Figure \ref{fig:alpha_ablation}. We can see that when $\alpha$ is set to $0.0$, only label entropy is used for calculating the reward. Its performance outperformed about 2.6 points compared with \text{AC-GAN+F}. When we only use smallest margin as the reward ($\alpha = 1.0$), the performance was boosted by 2.3 points. When $\alpha$ was set to $0.5$, the best performance was achieved, which is 0.5 and 0.8 points higher than those when $\alpha$ was set to $0.0$ and $1.0$, respectively. This implies that both criterion of smallest margin and label entropy are helpful for generating informative samples. 

\begin{figure}[!htp]
\centering
\centering
   \subfigure[AC-GAN]
  {\includegraphics[scale=0.31]{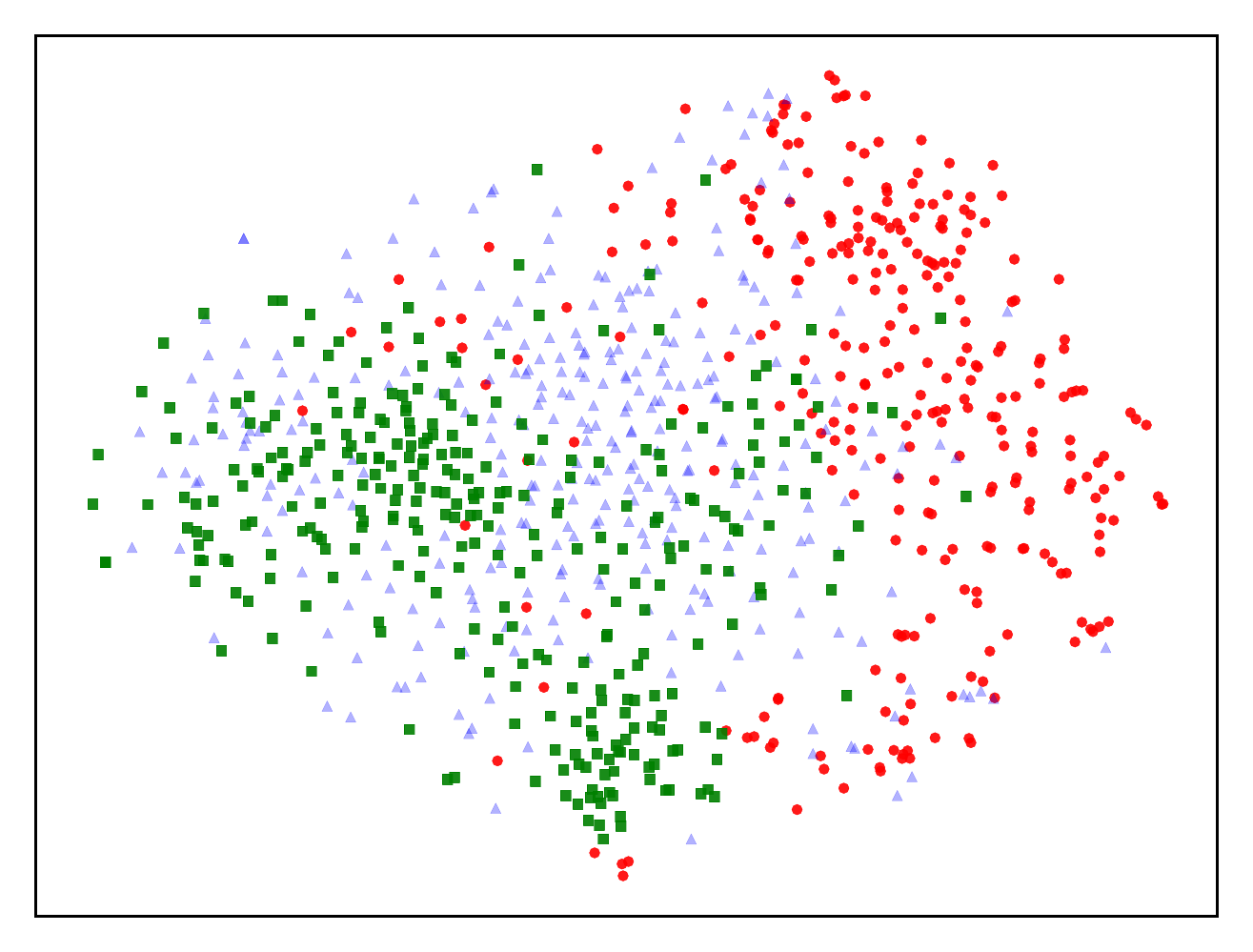}\label{subfig:ac-gan_dis}}
  \subfigure[ActiveGAN]
  {\includegraphics[scale=0.31]{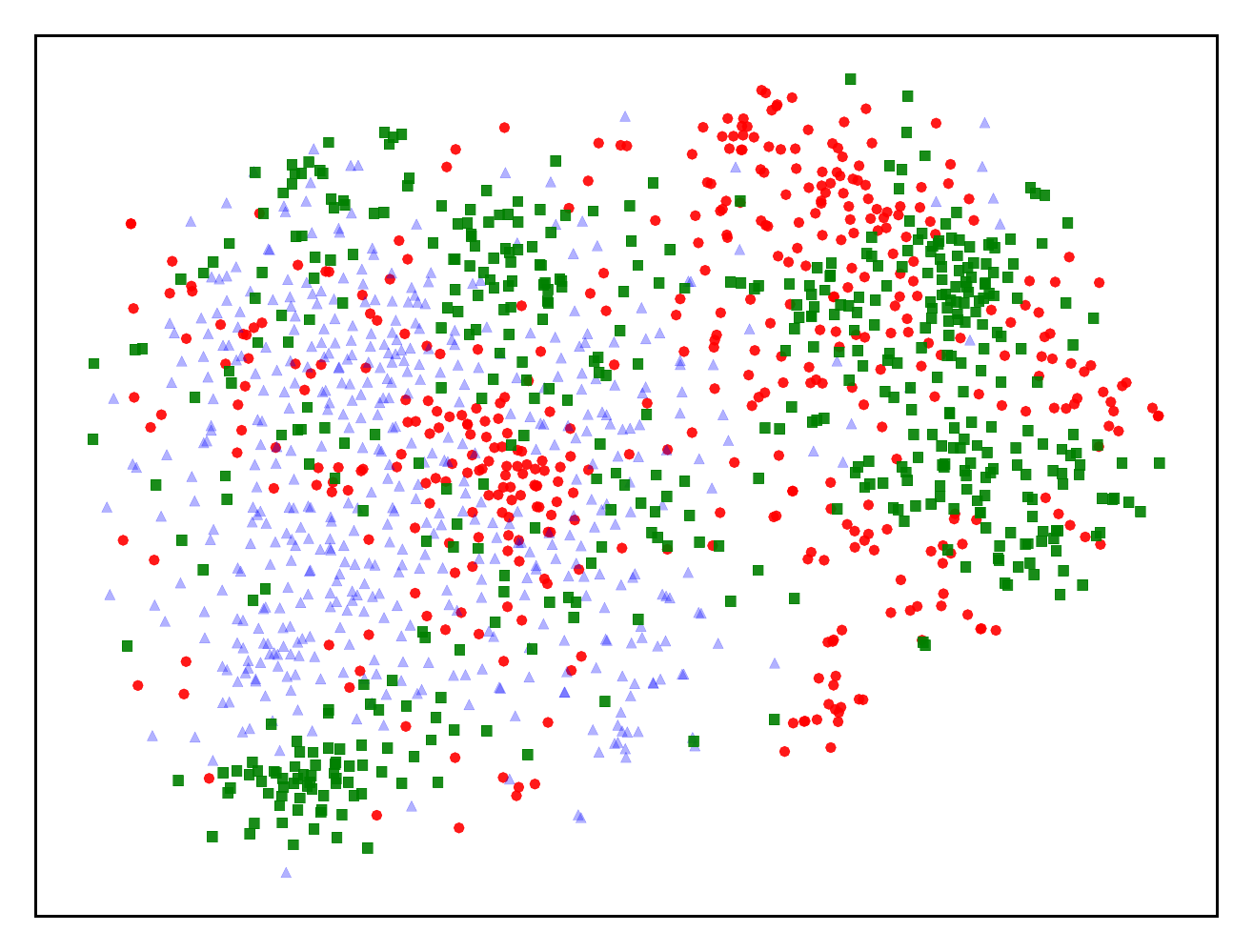}\label{subfig:active-gan_dis}}
\caption{This is t-SNE visualization of the generated samples from AC-GAN (a) and ActivGAN (b) on CIFAR-10, which are denoted by green points. The red points are the real hard samples selected from test data pool by using smallest margin, which is calculated by Equation \ref{eq:smallest_margin}. For a fair comparison, we set the smallest margin to $0.2$, which is the same as $\epsilon$ = 0.2 used in ActiveGAN. The light blue points denote training samples randomly sampled from each class.}
\label{fig:visual_feature_space}
\end{figure}

\begin{figure}[!htp]
\centering
\includegraphics[width=0.32\textwidth]{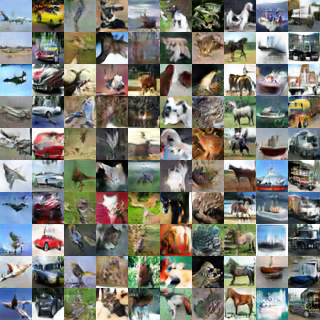}\label{fig:cm_cifar10_gal}
\caption{Samples of generated images from \text{ActiveGAN} in the dataset CIFAR-10. Each column shares the same label and each row shares the same latent variables.}
\label{fig:samples}
\end{figure}

To verify that \text{ActiveGAN} is more likely to generate informative images than \text{AC-GAN}, we visualized the features of training images and the generated images in CIFAR10 in a $2$-dimensional space, as shown in Figure \ref{fig:visual_feature_space}. The $2$-dimensional space is obtained by t-distributed stochastic neighbor embedding (t-SNE). As shown in Figure \ref{subfig:ac-gan_dis}, the generated samples of AC-GAN highly overlap the training samples distribution. As shown in Figure \ref{subfig:active-gan_dis}, by using our loss about uncertainty, the generated samples of ActiveGAN tends to distribute outside of the training samples. It can be also seen that some generated samples from ActiveGAN fall in the area of real hard samples, which is very unlike AC-GAN. 

We showed the sampled images from original training images and \text{ActiveGAN} on CIFAR-10, as depicted in Figure \ref{fig:samples}. The first column of Figure \ref{fig:samples} represents images of class label `airplane'. We can see that some images in Figure \ref{fig:samples} look like `bird', which might serve as informative images to discriminate the classes of `airplane' and `bird'.

\section{Conclusion}

In this paper, we investigate the problem of \textit{lack of labeled data}, in which labels of data can be obtained without using any additional unlabeled data nor querying the human oracle. In this work, we use class-conditional generative adversarial networks (GANs) to generate images, and devise a novel reward related to the degree of \textit{uncertainty} for generated samples. This reward is used to guide the class-conditional GAN to generate informative samples with a higher probability. Our empirical results on CIFAR10, MNIST, Fashion-MNIST and Tiny-ImageNet demonstrate that our proposed model is able to generate informative labeled images that are confirmed to be effective in improving classification performance. 

\bibliographystyle{aaai}
\bibliography{ebib}
\end{document}